# Emergent Behaviors in Multi-Agent Target Acquisition

Piyush K. Sharma[a], Erin Zaroukian[a], Derrik E. Asher[a], and Bryson Howell[b]

[a]DEVCOM Army Research Laboratory, Adelphi, MD 20783
[b]The University of Tennessee, Knoxville, TN 37996

## ABSTRACT

Only limited studies and superficial evaluations are available on agents' behaviors and roles within a *Multi-Agent System (MAS)*. We simulate a MAS using *Reinforcement Learning (RL)* in a *pursuit-evasion (a.k.a. predator-prey pursuit)* game, which shares task goals with *target acquisition*, and we create different adversarial scenarios by replacing RL-trained pursuers' policies with two distinct (non-RL) analytical strategies. Using heatmaps of agents' positions (state-space variable) over time, we are able to categorize an RL-trained evader's behaviors. The novelty of our approach entails the creation of an influential feature set that reveals underlying data regularities, which allow us to classify an agent's behavior. This classification may aid in catching the (enemy) targets by enabling us to identify and predict their behaviors, and when extended to pursuers, this approach towards identifying teammates' behavior may allow agents to coordinate more effectively.

**Keywords:** ISTAR, AI, MARL, MAS, Machine Learning, Reinforcement Learning, Predator-Prey, Pursuit-Evasion, Emergent Behavior, Coordination, Intelligence, Surveillance, Target Acquisition, Reconnaissance

## 1. INTRODUCTION

Future Army battlefields will carry out operations in complex and uncertain domains with emphasis on agility, perceptivity, resiliency, and reliability across a heterogeneous team of human-agent networks. Such symbiosis of human and robot Soldiers is likely to lead to performance that surpasses that of human Soldiers alone, even when equipped with advanced capabilities, such as increased speed, maneuverability, and lethality. Agents augmented with Artificial Intelligence (AI) can increase the survivability of human Soldiers and help eliminate threats on the battlefield. In order to maximize a Force's chance of winning by taking critical actions, agents must have a thorough training of the battlefield scenarios in simulated or actual environments. They should learn to use mounted tools effectively for target acquisition, and develop and sustain tactical skills to allow them to maneuver effectively on the battlefield.

The U.S. Army's *Multi-Domain operations (MDO)*[1] include several tasks including *intelligence, surveillance, target acquisition, and reconnaissance (ISTAR)*.[2] These sub-tasks can be represented in scenarios with simplified environments to gain an understanding for how *Multi-Agent Systems (MAS)* might perform when trained with *Reinforcement Learning (RL)* approaches in such operations. The simplified environment we use in this work is a continuous version of the *pursuit-evasion* task, which represents a *target acquisition* task with simple agent actions, where agents are able to continuously adjust their accelerations in a 2D plane. Examples of this type of target acquisition task include a scenario where ground robots pursue and attempt to maintain knowledge of an evader robot's position. In such a scenario, it would be of interest to show what behaviors a *black-box* method like *Multi-Agent Reinforcement Learning (MARL)* learns when gaining experiences through action-observation loops, where agents utilize a *centralized learning, decentralized execution (CLDE)* approach.

The target acquisition process requires a series of progressive and interdependent actions, which include observation (search), detection, location, identification, classification, and confirmation in a timely manner.[3] However, for unexplored terrains,[4] the search space can be very complex, making detection and identification tasks difficult for MAS. These tasks become even more challenging when pursuing an agile and self-aware target. Such MAS supposedly are enabled with AI capabilities that can coordinate and work together, forming a cohesive network to carry out autonomous intelligence, surveillance, and reconnaissance (ISR) tasks. Despite the success stories in military operations, such autonomous systems are far from being perfect and need to be appropriately

---

Corresponding author (Piyush K. Sharma)

evaluated before they can be deployed to defend critical territories. One example is Israel's *Iron Dome* defence system, which is claimed to have a success rate of *85% - 90%* in intercepting approaching missiles within a *2.5 - 43 mile* range.[5, 6] For systems like this to work successfully, detection and tracking radars need to coordinate with a firing system, and an interceptor must be able to maneuver quickly. Even with these prerequisites in place, this success rate may not be acceptable when a missile is targeting critical infrastructure like a bridge or supply route. Therefore, our dependency on an inadequately tuned AI algorithm may result in poor performance with an unacceptably high false negative rate, leading to mission failure. Moreover, agility, resiliency, and scalability in current MARL systems are some of the other challenges towards developing fully autonomous systems without requiring any human intervention.[7] Another typical challenge in developing such systems is scarcity of training data from actual combat zones to be able to realize realistic problem scenarios of a moving (potentially intelligent) target.[8–12]

In this paper, we set out to explore a MARL-based approach in a *pursuit-evasion* game, which shares task goals with *target acquisition*. We create different adversarial behaviors by replacing RL trained pursuers' policies with two distinct (non-RL) analytical strategies, namely, *Chaser* and *Interceptor* (see Section 2.2) and categorize an RL trained evader's behavior by creating an influential feature set. Our choice of MARL is motivated by a CLDE mechanism that is used in our analysis to control shared global information by parameterizing agents in the game simulation.[13, 14] When MARL methods such as the *Multi-Agent Deep Deterministic Policy Gradient (MADDPG)* algorithm[15] are used to train teams of agents in cooperative tasks, it has been observed that the actions of individual agents are significantly influenced by the actions of their teammates.[16, 17] Additionally, prior work has shown that teams of agents trained independently of one another under identical conditions display a variety of behaviors.[18] Moreover, these prior works have demonstrated the existence of coordination between teams of agents, indicating that MARL algorithms, such as MADDPG, are capable of producing emergent collaborative behaviors. If agents can identify these strategies, they can become more adaptive to new teammates by adjusting their behavior to accommodate their partner's strategy.

In order to work towards adaptable computational teammates, we have designed a method to identify the strategy employed by an agent within a *pursuit-evasion* game.[19] Specifically, we train a system of 4 MARL agents, 3 pursuers and 1 evader, to perform a pursuit-evasion task in an empty (obstacle-free) 2D bounded particle (simplified dynamics) simulation environment. We collect behavioral data and demonstrate that certain features are particularly useful for differentiating among agent behaviors. We verify that our method is capable of meaningfully describing differences in team strategies by testing it on teams of agents with well-defined strategies. We expect that this work will lay the ground work for future attempts to classify agent behaviors or team strategy.

This work proposes an approach for data preparation using attributes specific to evader dynamics in the pursuit-evasion game. Our histogram-based method bins an evader's locations across an entire episode. We evaluate how the choice of histogram bin size alters the identification of an agent's spatial patterns (i.e., evasion behavior). Specifically, we identify the bin size that is able to capture differences in RL trained evader's behavior when pursued by agents using different analytical strategies.

We provide visually discriminable patterns of the analytical strategies by finding a low dimensional representation of our data with *Principal Component Analysis (PCA)*.[20] We use PCA to explore the impact of state-space variables that allow us to distinctly separate analytical strategies playing against the same RL trained evader (i.e., all cases played the same evader policy). Furthermore, we show an emergent bimodal circling behavior of an RL-trained evader when pursued by a group of pursuers using analytical strategies. Our analysis indicates that an evader's learned policy allows it to successfully evade pursuers employing new and different behaviors. Our findings provide a foundational framework for future investigation aimed at deciphering the emergent behaviors from MARL agents.

In Section 2 we discuss our methodology to prepare the simulation environment for the *pursuit-evasion* game, provide a description of *analytical strategies*, and describe our dataset. In Section 3 we provide visual, quantitative, and classification results. In Section 4 we provide detailed discussion of our results. Finally, in Section 5 we provide conclusion and directions for future work.

## 2. METHODOLOGY AND EXPERIMENTAL DESIGN

The *pursuit-evasion* (aka *predator-prey pursuit*) game has recently been used to study coordination within MAS by exploring their interdependent action-selection choices. Although a vast literature exists on this game, many challenges still remain towards quantifying coordination among its cooperative agents. The pursuit-evasion game requires knowledge of the evader's (prey's/target's) location and so is aligned with an Army *Target Acquisition* (TA) task and part of the larger ISTAR mission objectives.

One challenge in analyzing a pursuit-evasion game in an Army context is to get relevant data. Moreover, a given dataset may result in a trained model that is specific to a particular environment and may not provide a sufficiently general solution. Therefore, given the lack of sufficient data for experimentation, we generate a simulation environment with controlled parameters to test various output scenarios. Specifically, we use this environment to demonstrate how a RL agent's trained policy (no longer learning) contains behaviors not demonstrated in test episodes with its RL trained adversaries to explore the potential for a RL policy to work in various untrained, unobserved, or novel situations.

### 2.1 Simulation Environment

An MAS in a continuous bounded 2D simulation environment was utilized to train and evaluate a set of 4 agents (3 homogeneous pursuer agents and 1 slightly advantaged evader agent) in the pursuit-evasion task with the MADDPG algorithm.[15] In training, pursuers received a shared positive reward every time one of them collided with the evader, and evaders received a corresponding negative reward. In test, pursuers received a point for each collision with the evader. Capabilities and environmental parameters match those from prior work.[16,18,21]

The simulation environment was built upon the *OpenAI Gym*[22] multi-agent particle environment repository and developed for use with MAS. A single model was trained with $100,000$ episodes at 25 time steps per episode. During the testing phase, data was collected per pursuer team configurations for 1000 episodes at 1000 time steps per episode. The pursuer team configurations were either analytical strategies (combinations of Chasers and Interceptors) or the originally trained RL policies (i.e., 3 RL pursuers). For all pursuer team configurations (or experimental conditions), the evader used its RL trained policy.

### 2.2 Analytical Strategies

In testing the MAS trained above, we substitute analytical strategies for trained policies. We define an analytical strategy as an explicit strategy that has been predefined using a simple formula to determine an agent's action. Pursuers' analytical strategies demonstrate distinct behaviors that are expected to contrast with behavior that emerges from pursuers' learned policies. This allows us to observe whether and how the evader's behavior (from a learned policy) is dependent on the behavior of the pursuers. Once we establish the level at which an RL evader's behavior is dependent on the pursuers' behaviors, we can use similar methods to determine if and how an RL purser's behavior is dependent on the behavior of its teammates. In this way, future work will allow us to assess cooperative as well as adversarial coordination.

In nature,[23] predators (pursuers) gradually try to approach their prey to decrease chase distance and time. However, this type of pursuit requires group coordination and speed, which may have varying degrees of success and may require a long engagement. Also, to intercept a target, a pursuer often benefits from stealth, such that the target does not alter its trajectory and the resulting point of intersection. Here we expect ambush predators to charge at target prey with very high speeds minimizing the distance between themselves and their prey over short time scales. This requires the predator to avoid detection, attain immediate accelerations to achieve desired speeds, and retain control over maneuver kinematics.

In our paper we are using a simplified version of the aforementioned natural strategies. We refer to pursuers who use the first strategy as *Chasers*, and our algorithm causes the Chaser to constantly attempt to minimize its distance from an Evader at each time step by assigning the Chaser an action that simply causes it to head toward the Evader's current location with maximum speed and acceleration. We refer to pursuers who use the second strategy as *Interceptors*, and our algorithm causes the Interceptor to head to the closest point at which it can intercept the evader based on the evader's current speed and heading. In Figure 1 we illustrate these behaviors. Note that, while MADDPG is likely to result in coordination among pursuers, these analytical strategies do not take into account the actions of the pursuers' teammates, and so will not result in actual coordination.

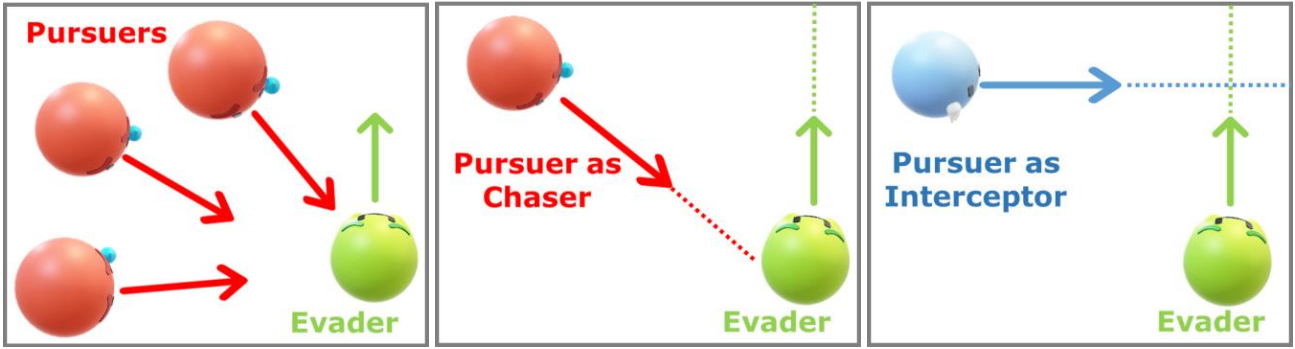

Figure 1: An illustration of **(Left)** pursuit-evasion game **(Middle)** Pursuer behaving as a *Chaser* **(Right)** Pursuer behaving as an *Interceptor*.

### 2.3 Pursuit-Evasion Data

In Section 2.2, we described that the testing was done by replacing pursuers' trained policies with analytical strategies. This created 4 different pursuer teams: *3 Interceptors*, *2 Interceptors + 1 Chaser*, *1 Interceptor + 2 Chasers*, and *3 Chasers*. An additional dataset was created using the pursuers' trained policies for comparison. These team configurations were each used to generate 1000 episodes of 1000 time steps each, resulting in a total of 5000 episodes across the 5 team configurations. These possible scenarios are summarized in Figure 2.

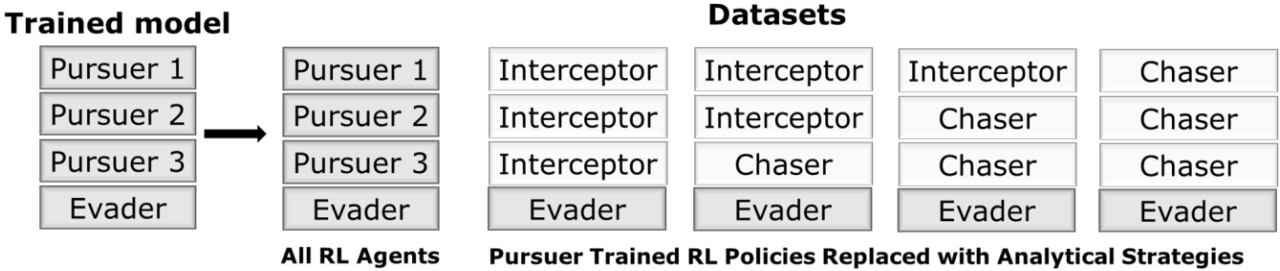

Figure 2: Description of training and testing with analytical strategies replacing pursuers' trained policies. Pursuer and evader trained policies are represented in gray rectangles, with chaser and interceptor analytical strategies in white rectangles. Note that the evader uses the same RL policy throughout.

For each of the 5000 episodes in our dataset, we created a 2D histogram of the target's location across the entire episode. These serve as heatmaps that indicate how many time steps the evader spent in different regions of space. Three versions of these histograms were constructed, each dividing the simulation environment into different granularities/quantities of bins: $5 \times 5$, $20 \times 20$, and $100 \times 100$. These histograms were then flattened into vectors of length 25, 400, and 10,000 respectively for analysis (Figure 3). Each vector represents an evader's spatial locations across all time steps in a given episode. Therefore, each dataset consists of 5000 vectors provided by each histogram granularity, and each of these 5000 vectors represents the evader's location for an episode in the 5 pursuer configurations (Figure 2).

### 3. RESULTS

We set out to explore the RL trained evader's policies when pursued by 3 agents using analytical strategies. In Section 2.3 we explained our approach for data preparation using varying bin sizes to get 2D histograms of the evader's location across the entire episode. In this section we see how the choice of a bin size impacts the identification of the agent's pattern. Specifically, we identify the bin size that is able to capture noticeable distinct patterns in RL trained evader's behavior when pursued by agents using different analytical strategies.

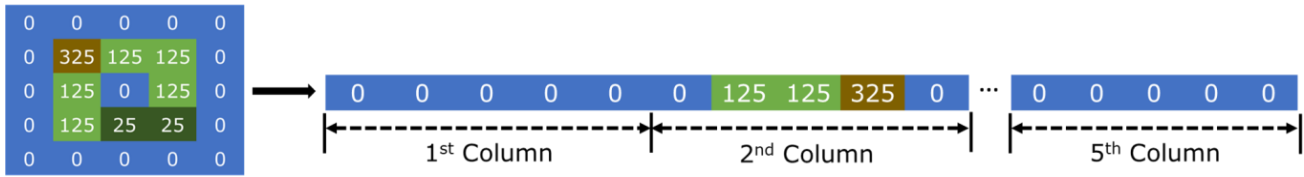

Figure 3: Illustration of a 5 × 5 2D histogram of a target's location in a single episode converted to a 1D vector of length 25.

## 3.1 Visualization of Emergent Behavior

For visually discriminating the RL trained evader behaviors in the 5 different pursuer team configurations, we explored various clustering and manifold techniques[20] on the spatial location data described in Figure 3. *Principal Component Analysis (PCA)* was utilized to identify a low-dimensional embedding of the data that allowed the separation of RL trained evader behaviors playing against 5 pursuer team configurations. PCA finds the low dimensional embedding by putting the transformed data to a new coordinate system such that the greatest variance by some projection comes to lie in the first few orthogonal principal components.[24]

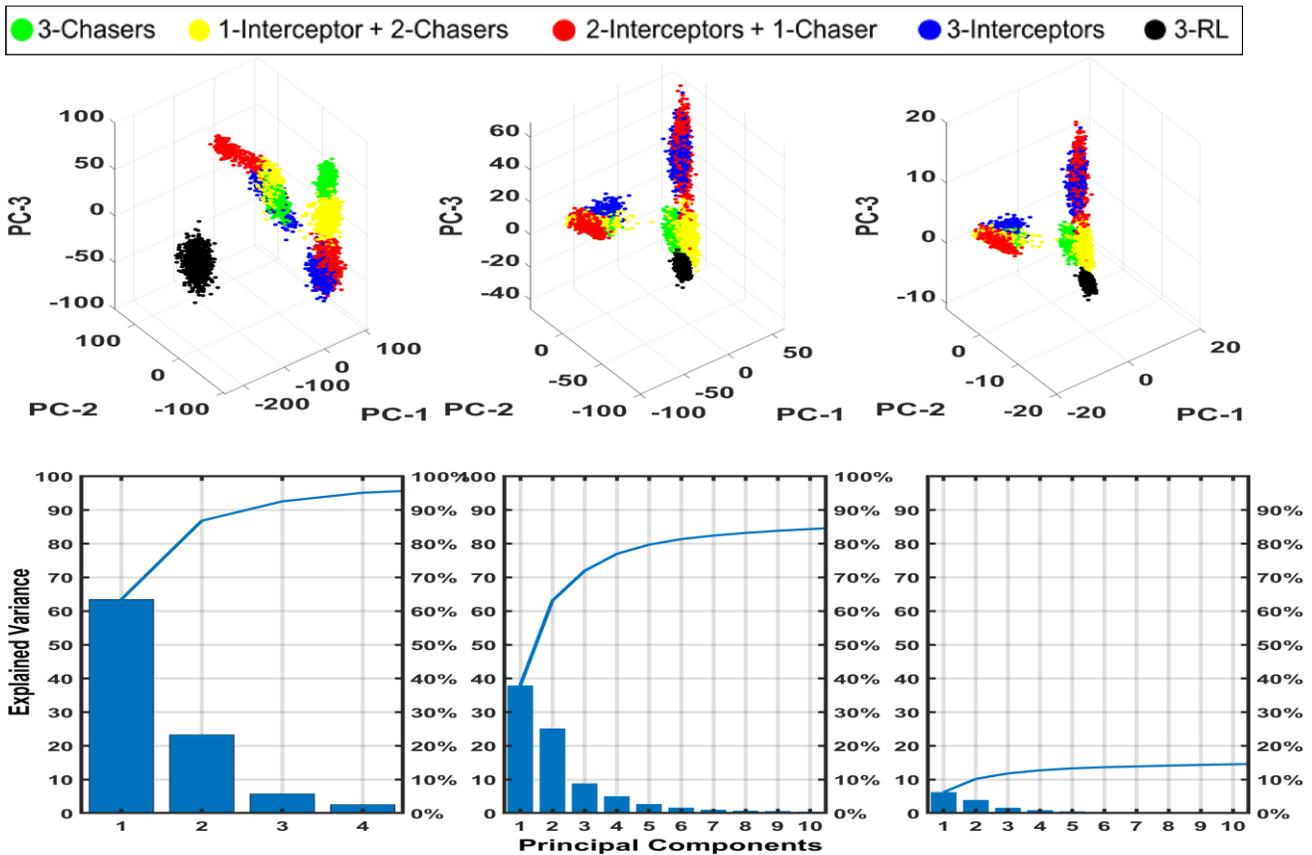

Figure 4: **(From Left to Right)** PCA results for bin sizes 5 × 5, 20 × 20, and 100 × 100 respectively. **(Top)** Scatter plots show that inter-cluster separability improves for smaller number of bins (i.e., larger bin size). RL trained agent (black) can be seen well-separated from other analytical strategy clusters (left column). **(Below)** Respective Pareto plots show explained variance decreases as the number of bins increases (i.e., smaller bin size).

3D scatter and *Pareto* plots were generated to visualize the differences in evader behavior given the policies of the pursuer team. The *Pareto* plots show the number of principal components calculated and variance explained per component, ordered from greatest to least variance explained per environment division (i.e., number of bins 5 × 5, 20 × 20, or 100 × 100). The cluster plots show the first three principal components for each environment division (5 × 5, 20 × 20, and 100 × 100) to illuminate the differences in evader behavior given the respective conditions (i.e., pursuer teams). From the scatter plots in Figure 4, smaller number of bins results in greater inter-cluster separability, and likewise, greater number of bins (i.e., finer-grained division of the environment) results in the RL trained pursuer team (black cluster) merging with one of the analytical strategy pursuer team clusters (compare black cluster in left cluster plot 5 × 5 to center 20 × 20 and right 100 × 100 cluster plots in Figure 4).

It can be seen that the best results were obtained for bin sizing corresponding to a 5 × 5 division of the simulation environment, with over 95% variance explained. The variance explained successively decreases for finer grained environment divisions 20 × 20 and 100 × 100, thus making the less granular environment division (i.e., fewest number of bins) an obvious choice for further analysis. In Section 3.2, the results are provided to illuminate how the fixed strategy pursuer teams elicited split clusters from the evader's circling behavior in Figure 6.

### 3.2 Quantitative Assessment of Analytical Strategies

To better understand the clustering results shown in Figure 4, a comparative analysis of the aforementioned analytical strategy pursuer team configurations was performed, based on pursuers' success rate with analytical strategy teams in Figure 5 (i.e., collisions with the evader), and the evader's resulting emergent circling behavior and mean circling radii (Figure 6). For comparison, the collisions, mean radius per episode, and evader's spatial locations are provided in Figure 7 for the baseline all-RL-agent case.

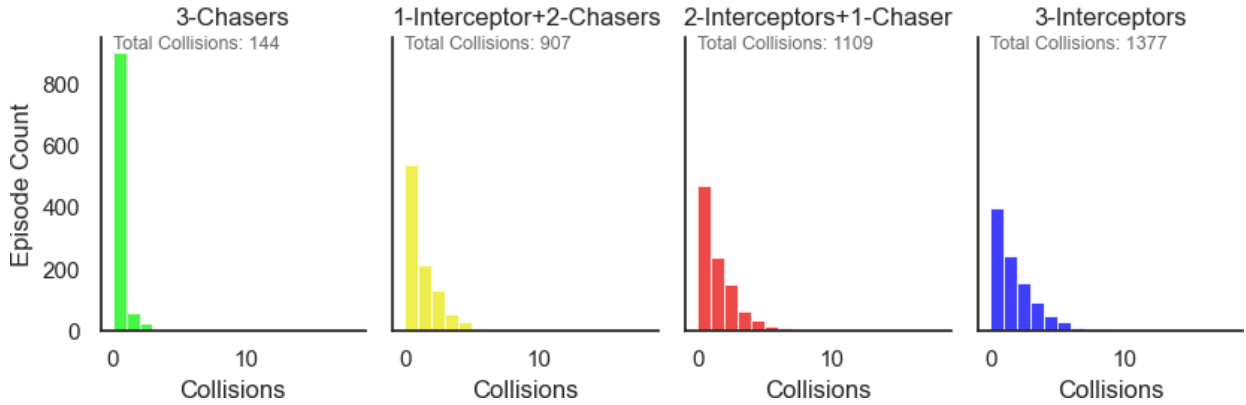

Figure 5: Histograms of the evader's collisions with the respective pursuer team configurations (conditions) from left to right:[3 Chasers], [1 Interceptor + 2 Chasers], [2 Interceptors + 1 Chaser], and [3 Interceptors]. The y-axes show the number of episodes in which a specific number of collisions occurred, with the x-axes showing the number of collisions. The total number of collisions across the 1000 test episodes is shown at the top of each plot.

A quick comparison between subplots in Figure 5 shows that the performance of the analytical strategy team configurations (or experimental conditions) increases with decreasing number of Chasers, where the 3 Chaser team performs the worst (144 collisions), with the 2 Chaser + 1 Interceptor team performing a little worse than the 2 Interceptor + 1 Chaser team (907 collisions compared to 1109 collisions) and the 3 Interceptor team performing the best with 1377 total collisions. Note that there is an order of magnitude difference in performance between the 3 Chaser and 3 Interceptor conditions (144 collisions versus 1377 collisions). However, it should also be noted that the performance of the RL trained pursuer team configuration performed nearly 2 orders of magnitude greater than the 3 Interceptor team (1377 collisions versus 99700 collisions). The total collision count for the 3 RL team condition is shown in Figure 7.

In contrast to the collision plots (Figure 5), the *Mean Radius* plots (Figure 6) reveal two distinct circling behaviors shown by the RL trained evader playing against all 4 of the analytical strategy pursuer team configurations. This distinct circling behavior should be compared to the much broader circling behavior that the same evader shows when playing against its originally trained adversaries (see 3 RL team in Figure 7). The mean radius for each of the 1000 episodes in the 4 analytical strategy configurations (or teams) was computed using Equation (1):

$$\frac{1}{n}\sum_{1}^{n}\left[\sqrt{(X_i - \bar{X})^2 + (Y - \bar{Y})^2}\right] \quad (1)$$

where *n* is the number of time steps, $X_i$, $Y_i$ are the coordinates of the evader's position at time step *i* and $\bar{X}, \bar{Y}$ are their respective mean values across all time steps in a given episode.

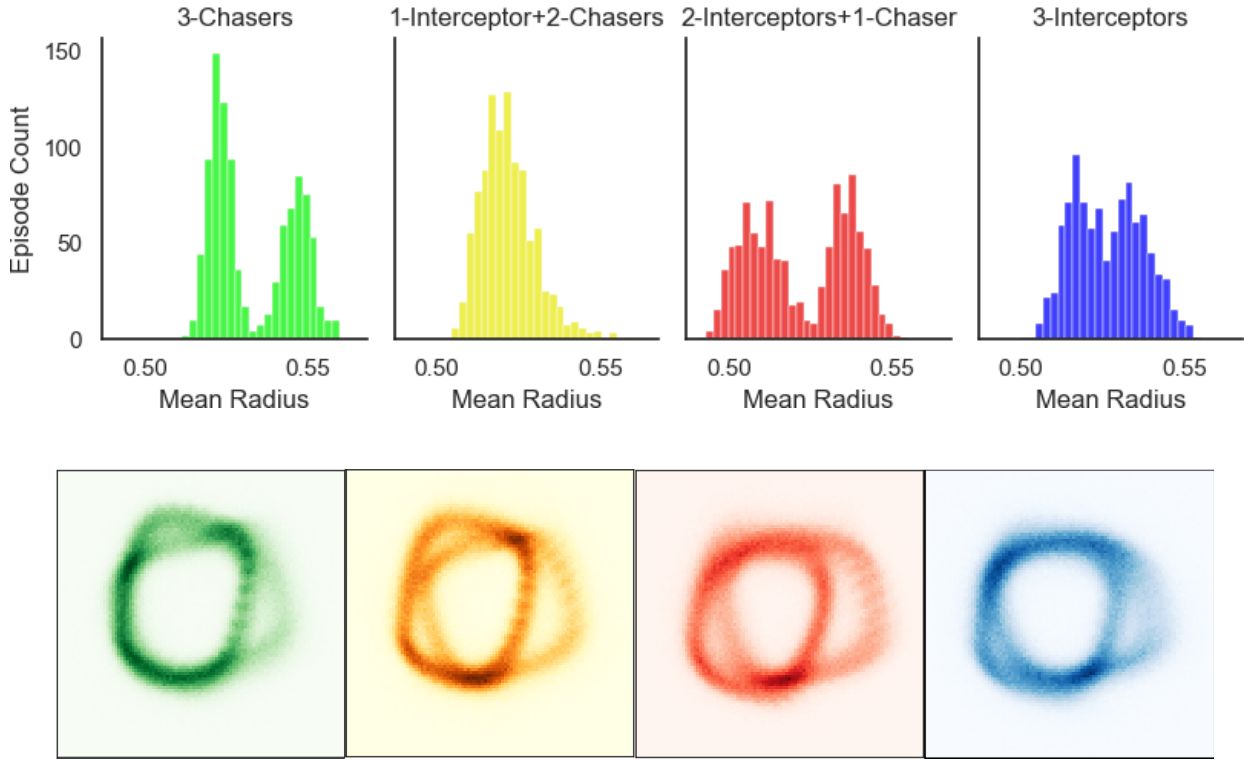

Figure 6: Evader's circling behavior is shown across the pursuer team configurations (conditions) per column. From left to right the conditions are: [green: 3 Chasers], [yellow: 1 Interceptor + 2 Chasers], [red: 2 Interceptors + 1 Chaser], and [blue: 3 Interceptors]. **(Above)** Histograms for the evader's *Mean Radius* across time steps per episode with the y-axes showing number of episodes and the x-axes showing the mean radius from the center of the environment for the evader's circling behavior. **(Below)** 100 × 100 heatmaps illustrating the evader's spatial locations across all 1000 time steps for each of the 1000 test data episodes for each of the 4 analytical strategy pursuer team configurations (conditions).

The mean radius histogram subplots shown in the top of Figure 6 indicate that the RL trained evader demonstrated a successful circling strategy (or perhaps pair of strategies) that was dramatically different from what is shown from training (compare top row of Figure 6 to center column of Figure 7). First, the histograms in the top of Figure 6 only contain mean radii between 0.5 and 0.55 (in arbitrary environment units ranging from -1 to +1 in x and y) across all 4 analytical strategy team configurations, whereas the RL trained evader playing against the original RL trained pursuer team is bounded between 0.75 and 0.85. Thus, the evader playing against the 4 analytical strategy pursuer team configurations produced circling behavior with smaller radii as well as less variance, as can be seen in the heatmaps in Figure 6 vs Figure 7.

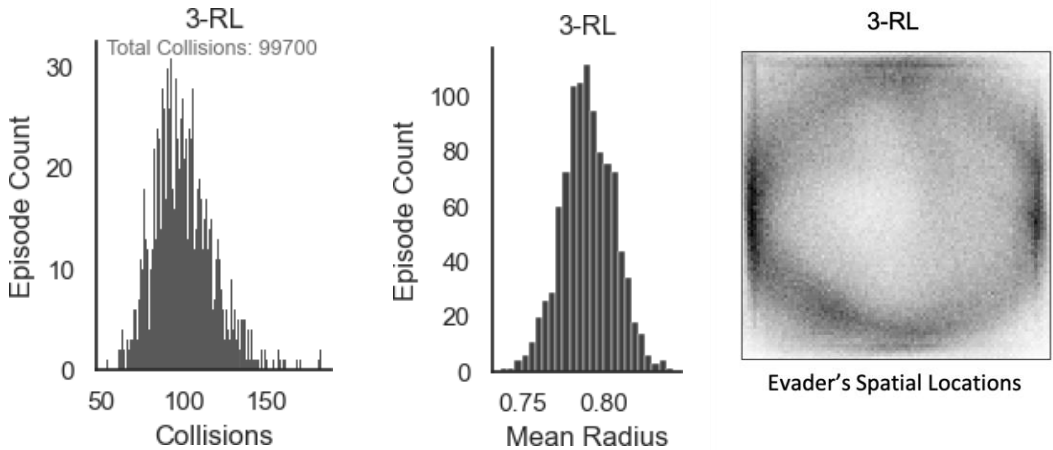

Figure 7: RL trained pursuer team configuration (condition) showing the evader's RL trained and tested behaviors. The histograms (left and center) are scaled differently from Figures 5 and 6 to illuminate the evader's originally trained behaviors. The total collisions are reported at the top of the left plot. The 100 × 100 2D histogram (right) shows a heatmap of the evader's spatial locations across the 1000 time steps per episode and all 1000 test episodes.

### 3.3 Classification

In Sections 2.2 and 2.3 we discussed the 4 configurations of analytical strategies for the 3 pursuing agents in a *pursuit-evasion* game. We used the 3 RL trained pursuers as a baseline for comparison (Figure 2). From Figure 6, it can be seen that subtle differences emerged in evader behavior across the 4 pursuer configurations; however, this result is difficult to discriminate visually in the heatmaps. Therefore, to further evaluate the differences in these cases, we introduced various *Machine Learning (ML)* algorithms to validate the classification of the spatial location data (demonstrated in Figure 3) that was observable through PCA and clustering in Section 2.3.

Given the broad range of ML algorithms available in the research literature, the choice for a suitable algorithm can be challenging. For this reason, we implemented several algorithms to show that the choice of algorithm did not have an impact on the validity of our results. A summary table of the results is shown in Table 1.

To find a suitable algorithm to verify the classification of the multiclass data presented in this work, we utilized the ML software WEKA[25, 26] and tuned the hyperparameters manually. Results were validated through 10-fold cross-validation on the randomized data sets constructed from *one-vs-one* classification problems drawn from the instances of the 5 pursuer team configurations (3 RL, 3 Chasers, 2 Chasers + 1 Interceptor, 1 Chaser + 2 Interceptors, and 3 Interceptors). We report the sample mean for *Accuracy*, *Precision*, *Recall*, area under the *Receiver Operating Characteristic curve (ROC)*, area under the *Precision-Recall curve (PRC)*, and *Root Mean Square Error (RMSE)*. In addition, Table 1 shows the time to build the model for each algorithm to compare resource efficiency of the approaches.

Support Vector Machines (SVM) classification was implemented with the use of a *Radial Basis Function (RBF)* kernel and WEKA's SMO optimization algorithm.[27] The other employed algorithms were *Simple Logistic*, *Logistic Model Tree*,[28] *Multilayer Perceptron*,[29] *Logistic Regression*,[30] *Random Forest*,[31] *J48*,[32] *Bagging*,[33] *Classification via Regression*,[34] and *Random Committee*.[35] Table 1 provides a summary of the classification results with Multilayer Perceptron highlighted in red to show it performed the best, while J48 performed the fastest. It is important to note that all algorithms exceeded 94% in accuracy, precision, recall, ROC, and PRC.

Table 1: Classification Summary

| Method | Accuracy | Precision | Recall | ROC | PRC | RMSE | Time to Build Model (Seconds) |
|---|---|---|---|---|---|---|---|
| SVM with RBF Kernel | 96.36% | 96.4% | 96.4% | 98.5% | 94.4% | 31.77% | 9.59 |
| Simple Logistic | 97.12% | 97.1% | 97.1% | 99.8% | 99.4% | 9.61% | 5.65 |
| Multilayer Perceptron | **98.66%** | **98.7%** | **98.7%** | **99.9%** | **99.7%** | **6.73%** | 31.05 |
| Logistic Regression | 96.96% | 97% | 97% | 99.7% | 99% | 9.99% | 5.99 |
| Random Forest | 98.54% | 98.5% | 98.5% | 99.9% | 99.9% | 8.45% | 1.92 |
| Logistic Model Tree | 97.82% | 97.8% | 97.8% | 99.8% | 99% | 8.4% | 29.11 |
| J48 | 96.16% | 96.2% | 96.2% | 98.1% | 94% | 12.05% | **0.32** |
| Bagging | 97.32% | 97.3% | 97.3% | 99.8% | 99.3% | 9.22% | 1.19 |
| Classification via Regression | 97.78% | 97.8% | 97.8% | 99.9% | 99.6% | 8.92% | 2.75 |
| Random Committee | **98.04%** | **98%** | **98%** | **99.8%** | **99.4%** | **8.91%** | **0.49** |

## 4. DISCUSSION

This research addressed the Army modernization priority to develop AI systems that can flexibly adapt to the human Soldier and the surrounding environment. Mission success will depend on sufficient coordination between AI systems and the Soldier, enabling AI systems with critical action selection capabilities that will provide the Soldier with improved safety, enhanced situational awareness, and general force multiplication. It is expected that such AI systems will be able to generalize specific learning experiences to entirely new situations by adhering to the military operations Observe, Orient, Decide, and Act (OODA) loop.

In general, it is expected that human Soldiers augmented with AI systems will have increased lethality and survivability in the future battles with technologically advanced adversaries. Therefore, current research efforts focus on training simulated agents in highly simplified and controlled environments to understand basic principles of multi-agent coordination, to better prepare for future battlefield scenarios. Specifically, we used a *pursuit-evasion* game to emulate an abstracted *target acquisition* task to demonstrate how RL agent policies allow the emergence of different behaviors under varying adversarial conditions. Further, different adversarial team configurations were adopted by replacing RL trained pursuer policies with two distinct (non-RL) analytical strategies, namely, *Chaser* and *Interceptor* (see Figure 1).

Our developed method to categorize differences in RL-trained evader behavior using a spatial histogram based approach, which divided the simulation environment into bins (with different granularities/quantities for comparisons) and used the bins as features for behavior classification. Heatmaps were used to visualize differences between evader behavior across pursuer team configurations (i.e., experimental conditions). Simply, the absolute positions of the evader proved to be an effective state-space variable for identifying and accurately classifying differences in the evader's behavior across conditions (see Table 1). PCA was used in conjunction with 3D cluster plotting to show how the spatial binning method produced visual differences in evader behavior across experimental conditions.

In the PCA analysis, the results showed that inter-cluster separability improved with fewer bins (i.e., larger bin size), which is a counter-intuitive result given that fewer differences in spatial location are captured by larger bins. In addition, the spatial binning method with PCA provided strong discrimination of the analytical strategies from the RL trained pursuer condition (see Figure 4, black clusters). A further comparison based on pursuers' success rate of capturing the evader (collisions) showed that among all tested analytical strategy combinations, the 3-Interceptor case (blue) was the most efficient one, with 1377 collisions. Moreover, the heatmaps of the evader's location reveal two different circling behaviors. This bimodal behavior likely corresponds to the two distinct spatial patterns uncovered by PCA in Figure 4. This suggests that the evader is adjusting its circling radius to evade the pursuers. Further, it appears that the RL-trained evader contained effective evasion strategies or behaviors in its learned policy that were not demonstrated in test trials with its trained pursuer team (i.e., 3-RL case). Therefore, this work provides evidence that an agent's RL trained policy may contain effective behaviors that cannot be demonstrated unless the environmental conditions (e.g., adversarial behaviors) are different from the training environment. Also, notice that all analytical strategies except 1-Interceptor + 2-Chaser (yellow) show similar patterns in Figure 6. This means that circling radii for these strategies are very similar and therefore difficult to discriminate. A further analysis explored various ML algorithms to improve the classification of all 5 scenarios given in Figure 2. Although all models perform extremely well, *Multilayer*

*Perceptron* performs the best with lowest RMSE. Also, in terms of time complexity, *J48* is the fastest taking only 0.32 seconds to build the model.

## 5. CONCLUSION

This work presents a novel method for evaluating the learned behavior of *black-box* MARL agents performing a pursuit-evasion (target acquisition) task. Our analysis showed an RL trained evader's emergent behavior when pursued by a group of pursuers using analytical strategies. The evader showed a bimodal circling behavior with subtle differences indicating that the evader's learned policy contains multiple strategies to deal with different adversaries. This is intriguing since the learned polices are very distinctly different from the cases with analytical strategies used by pursuers. Future work would require generating more data for further analysis to confirm our findings. Moreover, we would like to investigate agents' training policies generated with other MARL methods in the Army realistic scenarios.

## ACKNOWLEDGMENTS

This research work is supported by the DEVCOM Army Research Laboratory (ARL). The views and conclusions contained in this document are those of the authors and should not be interpreted as representing the official policies, either expressed or implied, of ARL or the U.S. government. The U.S. government is authorized to reproduce and distribute reprints for government purposes notwithstanding any copyright notation herein.